\documentclass[fleqn,10pt]{wlscirep}
\usepackage[utf8]{inputenc}
\usepackage[T1]{fontenc}
\usepackage{booktabs}
\usepackage{multirow}
\usepackage{graphicx}
\usepackage{amsmath}
\usepackage{hyperref}

\title{Three Roles, One Model: Role Orchestration at Inference Time to Close the Performance Gap Between Small and Large Agents}

\author[1]{S. Aaron McClendon}
\author[1]{Jorge Gallego-Feliciano}
\author[2]{Stavros Zervoudakis}
\author[2, *]{Antonios Saravanos}

\affil[1]{Aimpoint Digital Labs, Atlanta, GA, USA}
\affil[2]{New York University, New York, NY, USA}

\affil[*]{please direct correspondence to: Dr. Antonios Saravanos (saravanos@nyu.edu)}

\begin{abstract}
Large language model (LLM) agents show promise on realistic tool-use tasks, but deploying capable agents on modest hardware remains challenging. We study whether inference-time scaffolding alone, without any additional training compute, can improve the performance of a small model in complex multi-step environments. Operating on a single 24GB GPU, we evaluate Qwen3-8B on the AppWorld benchmark under both full-precision (FP16, 12K context) and 4-bit quantized (AWQ, 32K context) configurations. Without any intervention, the raw model achieves just 5.4\% (FP16) and 3.0\% (AWQ) task goal completion. Guided by a systematic failure mode analysis, we introduce a three-tier inference scaffolding pipeline that deploys the same frozen model in three distinct roles: (1) a summarization model that preserves critical artifacts (tokens, credentials, API responses) while compressing dialogue history; (2) the main agent model that reasons over the compressed context; and (3) an isolated correction model that reviews and revises the agent's code output without access to conversation history, breaking repetitive failure loops. Applied to the same unmodified model, this scaffolding yields 8.9\% (FP16) and 5.9\% (AWQ) task goal completion, roughly doubling performance in both settings, with particularly strong gains on difficulty-1 tasks (15.8\%$\to$26.3\% FP16; 5.3\%$\to$14.0\% AWQ). On full-precision inference, our scaffolded 8B model surpasses DeepSeek-Coder 33B Instruct (7.1\%) from the original AppWorld evaluation, demonstrating that structured inference-time interventions can make small models competitive with systems 4$\times$ their size. We formalize the approach as a scaffolded policy over a frozen base model, three invocations of the same weights with different conditioning, drawing connections to test-time compute scaling and action-space shaping in reinforcement learning. 
\end{abstract}

\begin{document}

\flushbottom
\maketitle
\thispagestyle{empty}

\section*{Introduction}

Tool-augmented language models are increasingly used as agents that can reason over intermediate observations, invoke external tools, and act in software environments \cite{li2024agent_survey,yao2023react,schick2023toolformer}. Their promise is especially clear for realistic digital tasks, where success depends not only on generating a plausible next action, but also on maintaining state across long trajectories, consulting documentation, handling authentication, recovering from execution errors, and adapting plans as new observations arrive.

While many recent benchmarks evaluate tool use or reasoning in controlled settings, they often abstract away the constraints that dominate real deployments. In practice, language model agents must operate under limited hardware budgets, maintain state across long interaction histories, handle authentication and API failures, and recover from execution errors. We refer to this setting as deployment realism, specifically, evaluating agents under the same resource constraints, execution dynamics, and failure modes they encounter in real-world software environments.

AppWorld \cite{trivedi2024appworld} was designed to capture this regime. It evaluates agents on realistic multi-step tasks spanning nine interconnected applications and 457 API endpoints, requiring persistent state management, correct API usage, and robust recovery from intermediate failures. This makes AppWorld not merely a tool-use benchmark, but a test of whether a language model can function as a persistent software agent under realistic deployment constraints.

Performance on AppWorld, however, remains strongly associated with model scale. In the original benchmark evaluation, GPT-4o achieves 48.8\% task goal completion on \texttt{test\_normal}, whereas smaller open-weight models lag substantially: DeepSeek-Coder 33B Instruct reaches 7.1\%, LLaMA-3-70B-Instruct 20.8\%, and Mistral 7B 0.0\% \cite{trivedi2024appworld}. This gap is practically important. Many academic labs and smaller organizations cannot fine-tune or serve frontier-scale systems, and instead must operate within the memory and compute budget of a single consumer-grade GPU. In that regime, methods that improve a frozen model at inference time, rather than through additional training, are especially attractive.

Prior work suggests that structured test-time computation can materially improve model behavior without updating parameters. Chain-of-thought prompting, self-consistency, and tree-of-thought reasoning show that additional inference-time structure can improve reasoning performance \cite{wei2022chain,wang2023selfconsistency,yao2023tree}. More recent work extends this idea to self-correction and memory-aware inference in multi-step settings. These results motivate a concrete question for agentic systems: can carefully designed inference-time scaffolding make a small model materially more effective in a realistic tool-use environment, even when the underlying weights remain fixed?

We investigate this question using Qwen3-8B \cite{qwen3} on AppWorld under a strict single-GPU deployment budget (24\,GB VRAM). We evaluate two configurations: full-precision FP16 with a truncated 12K-token context window, and 4-bit AWQ quantization with the full 32K context window \cite{lin2023awq}. Baseline performance is weak in both settings: the raw model achieves 5.4\% task goal completion in FP16 and 3.0\% in AWQ. A trajectory-level failure analysis shows that these failures are not explained by parameter count alone. Instead, the model frequently loses critical state, mishandles credentials, violates API schemas, and becomes trapped in repetitive correction loops after early mistakes.

Guided by this analysis, we design a three-tier inference scaffolding pipeline that deploys the same frozen model in three distinct roles. A summarization module compresses dialogue history while preserving critical artifacts such as tokens, credentials, intermediate API outputs, and error resolutions. A main agent reasons over the compressed context and proposes the next action. An isolated correction module then reviews and revises the proposed code using execution feedback and relevant API documentation, but without access to the full conversational history. The method is entirely training-free: it does not modify model weights, introduce additional supervision, or alter the benchmark environment. Instead, it changes the effective policy through structured conditioning and staged inference over the same base model.

This simple intervention yields consistent gains. The scaffolded system improves task goal completion from 5.4\% to 8.9\% in FP16 and from 3.0\% to 5.9\% in AWQ, with the largest gains on difficulty-1 tasks (15.8\%$\to$26.3\% in FP16; 5.3\%$\to$14.0\% in AWQ). Under full-precision inference, the scaffolded 8B model numerically exceeds the 7.1\% result reported for DeepSeek-Coder 33B Instruct on the same benchmark split \cite{trivedi2024appworld}, although this difference is not statistically significant at the 95\% level given the benchmark size ($N=168$). The broader implication is not that small models match frontier systems, but that a meaningful share of small-model underperformance on agentic benchmarks arises from mechanical failures that can be mitigated at inference time.

Our work makes three contributions. First, we provide an empirical characterization of failure modes for a small AppWorld agent, showing that authentication failures, planning errors, and API-schema mismatches dominate unsuccessful trajectories. Second, we present a diagnostic-first design methodology for inference-time scaffolding, in which each component is motivated by recurrent error patterns observed in baseline behavior. Third, we show that this scaffolded policy can substantially improve the effective performance of a frozen 8B model running on consumer-grade hardware, clarifying the role of inference-time structure as a practical lever for resource-constrained agent deployment.

The remainder of this paper is organized as follows. We first review related work on tool-using language model agents, reinforcement learning for agent improvement, and inference-time methods for self-correction and memory management. We then describe the experimental setup and formalize our scaffolded policy for AppWorld as an inference-time composition over a frozen base model. Next, we present our failure mode analysis and the design of the three-tier scaffolding pipeline. We then report the main empirical results, including comparisons with larger models, ablations, and shifts in failure mode distributions before and after scaffolding. Finally, we discuss the implications and limitations of our findings and conclude with directions for future work.

\section*{Related Work}

Tool-augmented language models have emerged as a central paradigm for building agents that can reason, act, and interact with external software systems \cite{li2024agent_survey,yao2023react,schick2023toolformer}. Early frameworks such as ReAct \cite{yao2023react} showed that interleaving reasoning traces with actions can improve sequential decision-making, while Toolformer \cite{schick2023toolformer} demonstrated that language models can learn to invoke external tools as part of their generation process. These ideas have since developed into a broader literature on LLM agents operating in realistic digital environments, where success depends not only on local reasoning quality but also on the ability to preserve state, recover from failed actions, use tools correctly, and adapt plans over long horizons. In closely related API-centric settings, API-Bank and Gorilla emphasize that robust tool use depends on API selection, documentation grounding, and schema-conformant argument construction \cite{li-etal-2023-api,patil2024gorilla}. AppWorld \cite{trivedi2024appworld} encapsulates these challenges in a particularly demanding benchmark: agents must coordinate across nine applications and 457 API endpoints while managing authentication, schema adherence, observation interpretation, and multi-step task execution. In that sense, AppWorld is not merely a tool-calling benchmark, but a test of whether a language model can function as a persistent software agent under realistic execution constraints.

Published results on AppWorld suggest that strong performance remains closely tied to model scale and additional optimization. In the original benchmark evaluation, frontier closed-weight systems substantially outperform smaller open-weight models, and even relatively large open models exhibit a considerable gap \cite{trivedi2024appworld}. More recent systems have narrowed that gap through reinforcement learning and other post-training procedures rather than through architectural simplicity or deployment efficiency alone \cite{chen2025loop,wang2025sage}. This matters because the regime we study is different from the one implicitly assumed by much of the recent literature. Many practical deployments do not have the compute budget to fine-tune large models, run extensive environment rollouts, or serve frontier systems at scale. Instead, they operate under severe hardware constraints, often on a single commodity GPU, and therefore require methods that improve agent behavior without changing model weights. Snell et al.\ \cite{snell2024scaling} provide direct empirical support for this direction, showing that compute-optimal inference-time scaling can outperform a model larger in FLOPs-matched evaluations and improve efficiency by more than $4\times$ over best-of-$N$ baselines, suggesting that the scale gap observed on benchmarks like AppWorld is not necessarily a fixed constraint but a function of how inference compute is deployed. Our work is motivated by this setting and asks a narrower but practically important question: how much agentic performance can be recovered from a frozen 8B model through inference-time structure alone?

One prominent line of work addresses the language-agent problem by optimizing the policy through training. On AppWorld, LOOP \cite{chen2025loop} trains a 32B model directly in the environment using a memory-efficient PPO variant and reports strong improvements over prompting-only baselines. SAGE \cite{wang2025sage} similarly combines reinforcement learning with a skill library and sequential rollouts to improve agent performance while reducing redundant interaction. More broadly, FireAct treats language-agent behavior itself as an object of fine-tuning rather than relying on prompting alone \cite{chen2023fireact}. Related work has explored broader strategies for strengthening language agents through training-based self-improvement, including self-questioning and self-navigation mechanisms that reduce reliance on manually curated supervision \cite{zhai2025agentevolver}, as well as more general analyses of reinforcement learning for reasoning-capable language models \cite{zhang2025survey}. Jiang et al. \cite{jiang2025lamer} further show that Meta-RL can induce more structured exploration behavior in language agents. Taken together, these studies show that additional optimization can materially improve interactive behavior, especially on long-horizon tasks where naive prompting underperforms. At the same time, they also assume access to rollout budgets, reward signals, and training compute that are often unavailable in the low-resource deployment regime. Our approach is complementary to this literature: rather than optimizing the parameters of the policy, we modify the effective policy by restructuring the inference procedure around the same frozen model.

A second and increasingly influential line of work improves model performance by scaling computation at test time rather than scaling parameters or retraining. Chain-of-thought prompting \cite{wei2022chain} showed that explicit intermediate reasoning can unlock capabilities that are otherwise latent in pretrained models, while self-consistency \cite{wang2023selfconsistency} demonstrated that aggregating multiple sampled reasoning paths can improve reliability. Tree-of-thought reasoning \cite{yao2023tree} extends this perspective by framing inference itself as a search problem over partial thoughts. These methods established a broader principle: the behavior of a language model is not determined only by its weights, but also by the structure of the computation wrapped around those weights at inference time. This perspective is particularly relevant for small models, where failures often arise not only from insufficient knowledge but also from brittle decoding trajectories, premature commitments, and poor recovery from earlier errors.

More recent work has applied this principle to self-correction and agentic planning. Bohnet et al. \cite{bohnet2025selfcritique} show that models can critique and refine their own plans without requiring an external verifier, achieving strong results on structured planning tasks. Self-Refine and CRITIC further show that models can iteratively critique and revise their own outputs, including with the aid of external tools and feedback \cite{madaan2023selfrefine,gou2024critic}. Reflexion \cite{shinn2023reflexion} establishes an earlier and closely related result. Agents can improve across interaction trials by converting execution feedback into natural language self-reflections stored in an episodic buffer, reinforcing better decision-making without any weight updates. In code-generation settings, Teaching Large Language Models to Self-Debug and Debug like a Human demonstrate that execution feedback can support stepwise code repair and runtime-grounded revision \cite{chen2024selfdebug,zhong-etal-2024-debug}. Our correction tier operationalizes a single-step version of this idea, using the same frozen model to review and revise its own code output conditioned on execution feedback rather than full trajectory history. SPIRAL \cite{zhang2025spiral} pushes this idea further by embedding a planner, simulator, and critic within a search loop for API-use environments, effectively treating tool-use planning as a test-time deliberation process. Recursive Language Models \cite{zhang2025rlm} use recursive self-invocation to handle long or compositional inputs, suggesting that repeated calls to the same model can substitute, to some extent, for larger monolithic context processing. Proxy-tuning \cite{liu2024proxy_tuning} approaches the frozen-model problem from a different angle by composing the outputs of tuned and untuned models at decoding time, again illustrating that substantial performance gains can sometimes be achieved without directly updating the base model. Our method belongs to this broader family of training-free inference-time interventions, but differs in both target setting and mechanism. We focus specifically on long-horizon software-agent execution by a small frozen model, and we operationalize inference-time structure through role specialization: the same weights are used as a summarizer, a main acting agent, and a corrector. This makes the method closer to a scaffolded policy than to a single prompting trick, because the effective behavior emerges from the composition of several specialized invocations rather than from one enlarged prompt.

This emphasis on composition also connects our work to a growing literature on memory and context management for long-horizon agents. A recurring difficulty in agent benchmarks is that performance often degrades not because the model cannot solve the task in principle, but because it loses access to the information needed to continue solving it. MemGPT \cite{packer2023memgpt} established an early and influential approach to this problem, introducing OS-inspired hierarchical memory management that pages information between in-context and external storage to give LLMs the appearance of an extended context window --- a conceptual precursor to the context compression and artifact preservation strategy in our summarization tier. Dynamic Cheatsheet \cite{suzgun2025dynamic_cheatsheet} addresses this by providing a persistent and evolving memory that accumulates useful strategies across inference steps. ACE \cite{zhang2025ace} similarly treats context as an evolving playbook and updates it incrementally to prevent context collapse. A-Mem \cite{xu2025amem} develops a dynamic indexing framework for agent memory inspired by Zettelkasten-style organization, while ReMe \cite{cao2025reme} demonstrates on AppWorld that explicit procedural memory can substantially improve the performance of Qwen3-8B. Alongside these memory-centric approaches, LongLLMLingua shows that long contexts can often be compressed while retaining task-relevant information, which is especially relevant to our summarization tier \cite{jiang-etal-2024-longllmlingua}. Related work on learning from early agent experience further suggests that interaction traces themselves can become a source of implicit world modeling and self-reflection, even without explicit reward-driven optimization \cite{zhang2025early_experience}. These approaches all underscore a common point: for language agents, memory is often not a peripheral engineering concern but a central determinant of competence. In our setting, the key requirement is not just generic memory, but artifact-preserving compression of executable state such as credentials, API outputs, pagination status, and previously resolved errors.

\section*{Experimental Setup}

As our central question concerns what can be gained from inference-time
structure alone, the experimental setup is designed to keep all other factors
fixed. We evaluate a single frozen 8B model on AppWorld in a realistic
resource-constrained setting and compare its raw agent behavior against the
same model equipped with our scaffolding pipeline. The subsections below
summarize the benchmark, metrics, and deployment configurations used for this
comparison.



\subsection*{Benchmark}
We report, task goal completion (the percentage of tasks for which the agent 
achieves the specified goal state) as our primary evaluation metric. Each task is evaluated with a single 
trial (i.e., pass@1). Results are reported by difficulty level (1--3) and in aggregate.

It should be mentioned that we do not report \textit{scenario goal completion}. Although this metric captures 
whether broader scenario-level constraints are satisfied, we found that smaller 
models rarely achieve full scenario completion, resulting in near-zero scores that are 
not especially informative. In addition, prior work on AppWorld does not consistently 
report this metric, which limits comparability. For these reasons, we therefore focus on task 
goal completion as the primary and most comparable measure of performance.

Furthermore, concurrent work on AppWorld \texttt{test\_normal} reports 
results under a pass@4 or Avg@4 protocol, where each task is attempted 
independently four times and success is measured either as the probability of 
at least one successful run (i.e., pass@4) or as the mean success rate across trials (i.e., Avg@4). 
For example, ReMe \cite{cao2025reme} reports a memoryless Qwen3-8B baseline of 
32.85\% pass@4 and 14.97\% Avg@4 on AppWorld, compared to our 5.4\% pass@1 
baseline. This difference reflects the evaluation protocol rather than an 
underlying performance discrepancy. Indeed, multiple independent trials can substantially inflate 
observed success rates on stochastic tasks. All results in this paper use 
greedy decoding (\texttt{temperature=0}) and single-trial evaluation, making 
our numbers directly comparable to the original AppWorld leaderboard results 
\cite{trivedi2024appworld} but not to multi-trial protocols without an appropriate 
conversion.

\subsection*{Hardware and Model Configurations}
All experiments were conducted on a single NVIDIA GPU with 24\,GB VRAM. We 
evaluate Qwen3-8B in two deployment configurations:
\begin{itemize}
    \item \textbf{FP16 (Full Precision):} Served via vLLM with a truncated 
    context window of 12,000 tokens, \texttt{max\_num\_seqs=4}, and 
    \texttt{gpu\_memory\_utilization=0.95}. This configuration sacrifices 
    context length for numerical precision.
    \item \textbf{AWQ (4-bit Quantized):} Using AWQ quantization to fit the 
    model with the full 32,768-token context window on the same hardware. This 
    configuration trades numberical precision for the ability to maintain longer 
    conversation histories.
\end{itemize}
Both configurations use greedy decoding (\texttt{temperature=0}), a fixed 
random seed of 100, and a maximum of 3,000 completion tokens per generation. 
The agent architecture follows the \texttt{simplified\_react\_code\_agent} implementation 
from the AppWorld codebase.

\section*{Problem Formulation}

We formalize the AppWorld agent task as a partially observable Markov decision 
process (POMDP) $\mathbb{M} = (\mathbb{S}, \mathbb{A}, \mathcal{T}, \mathcal{R}, 
\mathbb{O}, \Omega)$, where $\mathbb{S}$ is the environment state space (database 
contents, application states, authentication sessions), $\mathbb{A}$ is the action 
space (API calls with their arguments), $\mathcal{T}: \mathbb{S} \times \mathbb{A} 
\to \Delta(\mathbb{S})$ is the transition function, $\mathcal{R}: \mathbb{S} \times 
\mathbb{A} \to \{0, 1\}$ is the sparse task-completion reward (note in AppWorld, 
the metric for success can also be broken down by number of passed unit tests), 
$\mathbb{O}$ is the observation space (API responses, error messages), and 
$\Omega: \mathbb{S} \to \Delta(\mathbb{O})$ is the observation function.

At each timestep $t$, the agent maintains a history $h_t = (o_0, a_0, o_1, a_1, 
\ldots, o_t)$ and selects an action according to a policy $\pi_\theta(a_t \mid h_t)$, 
where $\theta$ denotes the frozen LLM parameters. The objective is to maximize the 
expected task completion reward:
\begin{equation}
    J(\pi_\theta) = \mathbb{E}_{\tau \sim \pi_\theta} \left[ \mathcal{R}(s_T, a_T) \right]
\end{equation}
where $\tau = (s_0, a_0, \ldots, s_T, a_T)$ is a trajectory and $T$ is the 
terminal timestep.


\paragraph{Challenges for Small Models.}
In standard RL approaches, one would optimize $J$ by updating $\theta$ via policy gradient methods. However, under our constraint of no additional training compute, $\theta$ is fixed. The effective policy is further degraded by two practical limitations. First, the history $h_t$ must be truncated to fit within the model's context window $C$, so the agent operates on $\tilde{h}_t = \textsc{Trunc}(h_t, C)$ rather than the full history---inducing a form of partial observability beyond what the environment imposes. Second, quantization introduces an approximation $\hat{\theta} \approx \theta$ that degrades the policy's per-step decision quality.

\paragraph{Scaffolded Policy.}
Rather than optimizing $\theta$, we construct a \emph{scaffolded policy} $\pi^{\text{scaf}}$ that deploys the same frozen model in three distinct roles---each conditioned on a different information set---and composes their outputs. Concretely:
\begin{equation}
    \pi^{\text{scaf}}(a_t \mid h_t) = \underbrace{\pi_\theta^{\text{correct}}(a_t' \mid a_t, d_t)}_{\text{corrector}} \circ \underbrace{\pi_\theta^{\text{agent}}\big(a_t \mid \tilde{h}_t, o_t\big)}_{\text{main agent}} \quad\text{where}\quad \tilde{h}_t = \underbrace{\pi_\theta^{\text{sum}}(h_t)}_{\text{summarizer}}
    \label{eq:scaffolded_policy}
\end{equation}
The three model invocations share identical frozen weights $\theta$ but differ in their conditioning:
\begin{itemize}
    \item $\pi_\theta^{\text{agent}}$: the \textbf{main agent policy}, which receives the summarized history $\tilde{h}_t$ and current observation $o_t$, and proposes an action $a_t$.
    \item $\pi_\theta^{\text{sum}}: \mathcal{H} \to \mathcal{H}'$: the \textbf{summarization model}, which compresses the history $h_t$ into a shorter but informationally richer representation $\tilde{h}_t$, satisfying $|\tilde{h}_t| \leq C$ while preserving critical state variables (credentials, API schemas, error patterns).
    \item $\pi_\theta^{\text{correct}}: \mathcal{A} \to \mathcal{A}$: the \textbf{correction model}, which post-processes the agent's proposed action, operating on the action and relevant API documentation $d_t$ \emph{without} access to $h_t$:
    \begin{equation}
        \pi_\theta^{\text{correct}}(a_t') = \pi_\theta(a_t' \mid a_t, d_t)
        \label{eq:correction}
    \end{equation}
\end{itemize}

\begin{figure}[h]

  \centering
  \includegraphics[width=0.5\linewidth]{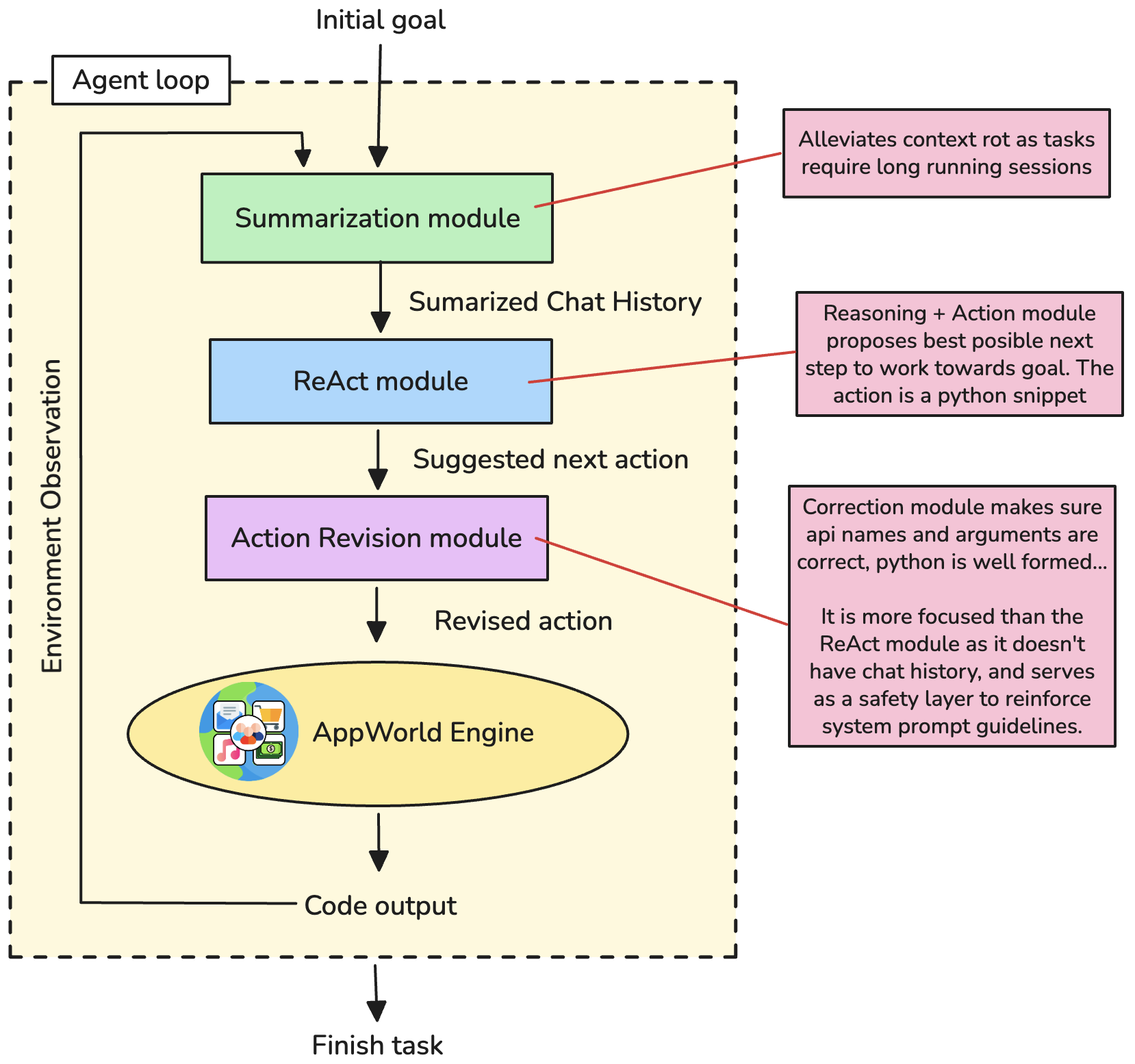}
  \label{fig:failmodes-diff-1}
  \caption{Our proposed modular architecture. Each module targets an important failure mode for small language models to increase performance.}
\end{figure}

The key insight is that a single frozen model, invoked three times with different contexts, can serve as its own critic and memory manager. The correction model (Eq.~\ref{eq:correction}) is deliberately denied access to $h_t$, which serves as an implicit \emph{regularizer}: by preventing the corrector from conditioning on the main agent's potentially corrupted reasoning history, we reduce the probability of perpetuating error cascades and repetitive loops identified in our failure analysis.

This formulation makes explicit that our approach modifies the \emph{effective policy} without modifying $\theta$---analogous to how reward shaping or action masking can improve RL policy performance without retraining, but applied at the level of input/output transformations around a frozen language model. The additional inference cost is modest: each agent step requires at most three forward passes through the same 8B model, all served on the same single GPU.

\section*{Method}

\subsection*{Failure Mode Analysis}
Before designing interventions, we conducted a systematic failure mode analysis on baseline outputs. Using GPT-4o as an evaluator, we classified each failed task into primary and secondary failure categories. The analysis covered all failed tasks across both model configurations in the Appworld train dataset split.

Table~\ref{tab:failure_modes} presents the primary failure mode distribution for the FP16 baseline. Three categories dominate: authentication/credential failures (28.2\%), reasoning/planning errors (26.4\%), and API parameter/schema mismatches (17.8\%). Together these account for over 72\% of all failures. Confidence-weighted analysis (where GPT-4o provides a confidence score for each classification) confirms this ranking, with authentication issues scoring highest at 39.45 weighted failure points.

Qualitative inspection of failure traces reveals that many of these categories are causally linked through a common failure cascade. A major breaking point is the model's tendency to call API endpoints without first consulting their documentation---despite explicit instructions in the system prompt to do so. This leads to parameter name or type mismatches on the initial call, which produces an error response. Rather than consulting the API documentation at this point, the model frequently attempts to brute-force the correct parameters through trial and error, consuming turns and context budget. When these repeated failed calls exhaust sufficient context, the model loses track of previously obtained authentication tokens and begins re-extracting credentials, entering a degenerative loop. Thus, what appears in the failure taxonomy as distinct categories (API schema mismatches, authentication failures, and repetition loops) often represents a single cascading failure mode originating from the model's reluctance to consult documentation proactively.

\begin{table}[ht]
\centering
\caption{Primary failure mode distribution for FP16 baseline on AppWorld \texttt{test\_normal} (163 failed tasks). Categories are ranked by frequency. Confidence-weighted scores incorporate the evaluator's classification confidence.}
\label{tab:failure_modes}
\begin{tabular}{lcc}
\toprule
\textbf{Failure Category} & \textbf{Count} & \textbf{Conf.-Weighted} \\
\midrule
Authentication / credential issue & 46 & 39.45 \\
Reasoning / planning error & 43 & 35.30 \\
Wrong API params / schema mismatch & 29 & 25.05 \\
Other & 24 & 20.05 \\
Missing API call / wrong API name & 13 & 11.45 \\
Repetition / loop & 3 & 2.55 \\
Formatting / code block error & 2 & 1.70 \\
Pagination / incomplete iteration & 2 & 1.65 \\
Context length / token limit & 1 & 0.80 \\
\bottomrule
\end{tabular}
\end{table}

Notably, the failure distribution shifts across difficulty levels (Table~\ref{tab:failure_by_diff}). Authentication failures become increasingly dominant at higher difficulties (22 of 62 difficulty-3 failures), while reasoning errors maintain a steady presence across all levels. We attribute this to the longer interaction sequences required by difficulty-3 tasks: the model typically extracts authentication tokens successfully in early turns, but as the conversation grows, these credentials are displaced from the effective context window. The model then enters a degenerative cycle of re-extracting credentials it has already obtained, consuming additional turns and further degrading context quality. This credential-loss loop is a direct consequence of the small model's limited context capacity under multi-step task pressure.

\begin{table}[ht]
\centering
\caption{Top-3 primary failure modes by difficulty level (FP16 baseline).}
\label{tab:failure_by_diff}
\begin{tabular}{clc}
\toprule
\textbf{Difficulty} & \textbf{Failure Category} & \textbf{Count} \\
\midrule
\multirow{3}{*}{1} & Auth / credentials & 13 \\
                    & Reasoning / planning & 11 \\
                    & Other & 11 \\
\midrule
\multirow{3}{*}{2} & Reasoning / planning & 13 \\
                    & API params / schema & 12 \\
                    & Auth / credentials & 11 \\
\midrule
\multirow{3}{*}{3} & Auth / credentials & 22 \\
                    & Reasoning / planning & 19 \\
                    & Other & 7 \\
\bottomrule
\end{tabular}
\end{table}

\subsection*{Three-Tier Inference Scaffolding}

Informed by the failure analysis, we instantiate the scaffolded policy $\pi^{\text{scaf}}$ (Eq.~\ref{eq:scaffolded_policy}) as three concurrent invocations of the same frozen Qwen3-8B model, each conditioned on a different information set and serving a distinct role. The data flow for each agent turn is: AppWorld observation $o_t$ $\to$ $\pi_\theta^{\text{sum}}$ (compress history) $\to$ $\pi_\theta^{\text{agent}}$ (generate action) $\to$ $\pi_\theta^{\text{correct}}$ (revise action) $\to$ output to AppWorld.

\paragraph{Tier 1: Main Agent Model ($\pi_\theta^{\text{agent}}$).}
The main agent receives the current observation $o_t$ from AppWorld along with the compressed history $\tilde{h}_t$ produced by the summarizer, and generates a proposed action $a_t$ (a Python code block containing API calls). This is the standard ReAct-style agent loop, operating within the \texttt{simplified\_react\_code\_agent} framework from the AppWorld codebase.

\paragraph{Tier 2: Summarization Model ($\pi_\theta^{\text{sum}}$).}
The summarization model addresses the context degradation problem identified in our failure analysis. It is triggered when the history length exceeds predefined thresholds ($|h_t| > 24{,}000$ characters or $> 6{,}000$ tokens), compressing the conversation while explicitly preserving the critical state variables that the baseline policy loses over long trajectories: Authentication tokens and credentials obtained during execution; API endpoint names and their observed response schemas; Error patterns and their resolutions; Pagination states and iteration progress; task completion status indicators.

The module retains the first $N=26$ and last $K=6$ messages verbatim, summarizing the intermediate history. Previously extracted artifacts (access tokens, API outputs) are required to be returned verbatim in the summary. The summarization model runs on a separate vLLM endpoint but uses the same frozen weights $\theta$. In POMDP terms, $\pi_\theta^{\text{sum}}$ reduces the induced partial observability from context truncation by ensuring that the compressed history $\tilde{h}_t$ retains higher mutual information with the latent environment state $s_t$ than naive truncation would.

\paragraph{Tier 3: Correction Model ($\pi_\theta^{\text{correct}}$).}
After the main agent generates a proposed action $a_t$, the correction model (Eq.~\ref{eq:correction}) intercepts it before submission to AppWorld. As formalized above, this model operates on $a_t$ and relevant API documentation $d_t$ \emph{without} access to the history $h_t$. It receives only: the agent's proposed code output; relevant API documentation for any endpoints referenced in the code; the most recent execution result (if a prior step failed).

The correction model is parameterized by the same frozen $\theta$ but prompted to classify the failure into a predefined error taxonomy, cite evidence from the execution output, provide a concise diagnosis, and output a corrected executable Python patch. The patch must contain exactly one code block, execute directly in AppWorld, include at least one \texttt{apis.*} call, and match API documentation argument names exactly. If the fix is uncertain, the model must minimally query API documentation via \texttt{apis.api\_docs.show\_api\_doc(...)}.

The history isolation serves as an implicit regularizer on the correction policy: by conditioning $\pi_\theta^{\text{correct}}$ only on $(a_t, d_t)$ and denying access to $h_t$, we prevent the corrector from inheriting the main agent's potentially corrupted state representation. This breaks the failure cascades identified in our analysis---where the policy conditions on its own erroneous reasoning chains and enters repetitive loops---by forcing corrections to be grounded solely in API documentation.


\subsection*{Limitations of the Correction Model at Higher Difficulties}
We observe that for difficulty-3 tasks, the regularization effect of history isolation in $\pi_\theta^{\text{correct}}$ can become a liability. Without access to $h_t$, the correction model may hallucinate placeholder values for state variables (e.g., access tokens, user IDs) that were established in earlier trajectory steps. In RL terms, the correction policy's observation set $(a_t, d_t)$ is insufficient for reconstructing the relevant latent state when tasks require long-horizon dependencies.

This limitation highlights a broader trade-off between robustness and state awareness: while restricting the correction model's inputs improves stability and reduces overfitting to trajectory noise, it also removes information necessary for tasks with extended temporal dependencies. Addressing this tension likely requires more principled mechanisms for state summarization or memory, rather than direct exposure to full trajectory history.

\section*{Results}

\subsection*{Main Results}

\begin{figure}[h]

  \centering
  \includegraphics[width=\linewidth]{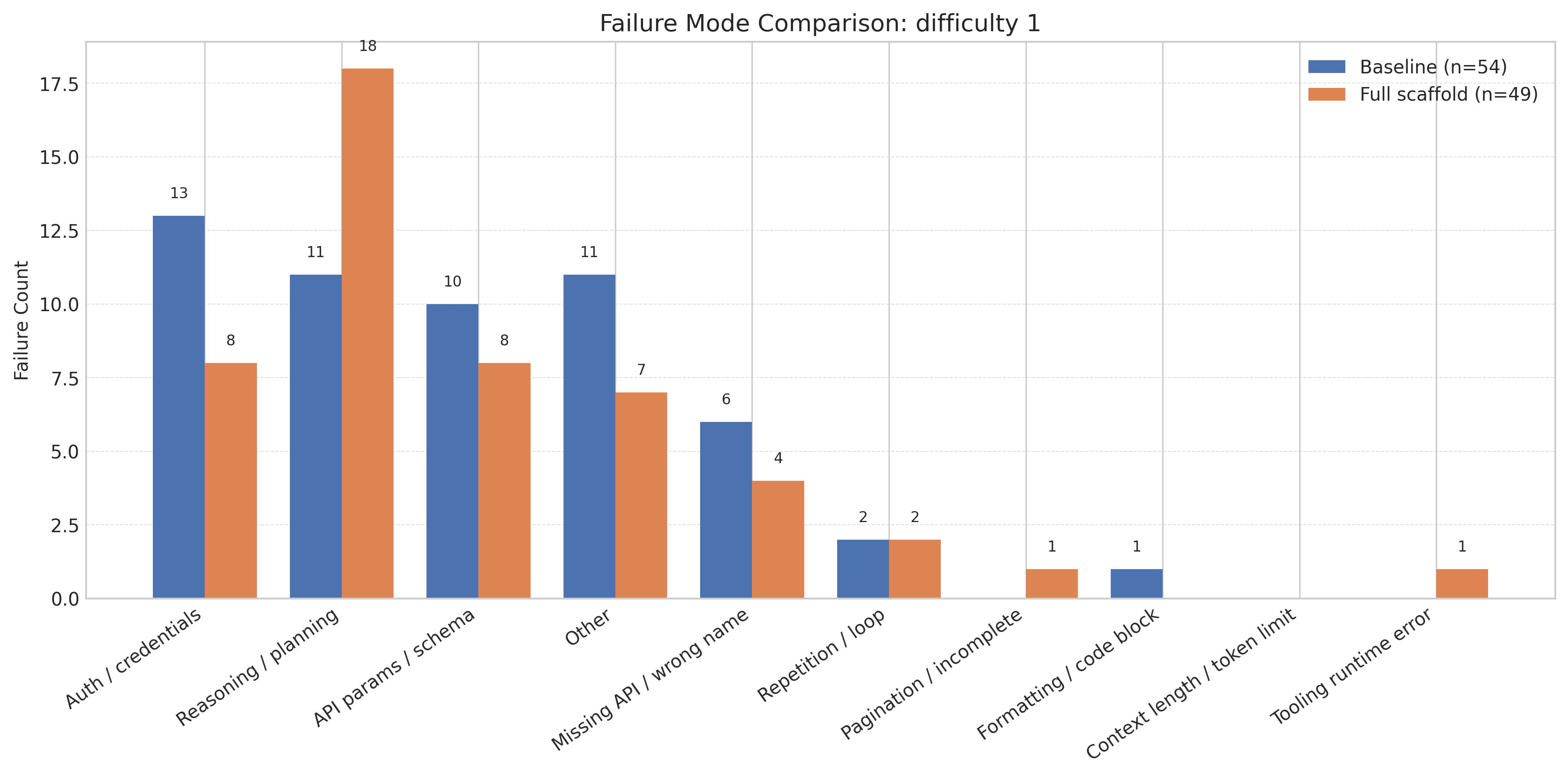}
  \label{fig:failmodes-diff-1}
  \caption{Comparison of failure modes for tasks of difficulty 1 for baseline vs full scaffold for Qwen 3 8B AWQ, which is a quantized version of Qwen.}
\end{figure}

Table~\ref{tab:main_results} presents our main results across all configurations. The inference scaffolding approximately doubles task goal completion in both the FP16 and AWQ settings. We report 95\% Wilson confidence intervals for all task goal completion rates.

\begin{table}[ht]
\centering
\caption{Task goal completion (\%) on AppWorld \texttt{test\_normal} (168 tasks). FP16 = full precision with 12K context; AWQ = 4-bit quantized with 32K context. “Scaffold” denotes our three-tier inference scaffolding method. We report 95\% Wilson confidence intervals for aggregate task goal completion rates.}
\label{tab:main_results}
\begin{tabular}{lcccc}
\toprule
 & \multicolumn{2}{c}{\textbf{FP16 (12K ctx)}} & \multicolumn{2}{c}{\textbf{AWQ (32K ctx)}} \\
\cmidrule(lr){2-3} \cmidrule(lr){4-5}
\textbf{Difficulty} & \textbf{Baseline} & \textbf{Scaffold} & \textbf{Baseline} & \textbf{Scaffold} \\
\midrule
Aggregate & 5.4 [2.8, 9.9] & \textbf{8.9 [5.5, 14.2]} & 3.0 [1.3, 6.8] & \textbf{5.9 [3.3, 10.6]} \\
Difficulty 1 & 15.8 & \textbf{26.3} & 5.3 & \textbf{14.0} \\
Difficulty 2 & 0.0 & 0.0 & 2.1 & \textbf{4.2} \\
Difficulty 3 & 0.0 & 0.0 & 1.6 & 0.0 \\
\bottomrule
\end{tabular}
\end{table}


Several patterns emerge. First, the scaffolding provides the largest absolute gains on difficulty-1 tasks: +10.5 percentage points for FP16 and +8.7 for AWQ. This is expected, as simpler tasks are more amenable to single-step corrections---the types of errors the correction node is best equipped to fix. Second, the FP16 configuration consistently outperforms AWQ at equivalent scaffolding levels despite having less than half the context window, suggesting that for this model and task distribution, numerical precision is more valuable than extended context. Third, difficulty-3 tasks remain largely intractable for the 8B model regardless of configuration, likely due to the multi-step reasoning chains that exceed the model's planning capacity.

\subsection*{Comparison with Larger Models}

Table~\ref{tab:comparison} contextualizes our results against published AppWorld numbers from the original evaluation \cite{trivedi2024appworld}. Notably, the comparison with DeepSeek-Coder 33B Instruct is direct: both evaluations use the same benchmark version, task split, and evaluation protocol. Our scaffolded FP16 configuration (8.9\%) surpasses the reported DeepSeek-Coder 33B Instruct score (7.1\%), a model with approximately 4$\times$ more parameters, under identical evaluation conditions, though the difference is not statistically significant at the 95

\begin{table}[ht]
\centering
\caption{Comparison with published AppWorld results (task goal completion, \texttt{test\_normal}). Our results use Qwen3-8B on a single 24\,GB GPU. For our models, and for DeepSeek-Coder 33B Instruct assuming the same 168-task evaluation set, we report 95\% Wilson confidence intervals.}
\label{tab:comparison}
\begin{tabular}{lccc}
\toprule
\textbf{Model} & \textbf{Parameters} & \textbf{task goal completion (\%)} & \textbf{95\% CI} \\
\midrule
GPT-4o & --- & 48.8& --- \\
Claude 3.5 Sonnet & --- & 33.2 & --- \\
LLaMA-3-70B-Instruct & 70B & 20.8 & --- \\
\midrule
\textbf{Qwen3-8B + Scaffold (FP16)} & \textbf{8B} & \textbf{8.9} & [5.5, 14.2] \\
DeepSeek-Coder 33B Instruct & 33B & 7.1 & [4.1, 12.0] \\
Qwen3-8B Baseline (FP16) & 8B & 5.4 & [2.8, 9.9] \\
\textbf{Qwen3-8B + Scaffold (AWQ)} & \textbf{8B} & \textbf{5.9} & [3.3, 10.6] \\
Qwen3-8B Baseline (AWQ) & \textbf{8B} & 3.0 & [1.3, 6.8] \\
\bottomrule
\end{tabular}
\end{table}


\subsection*{Ablation Study}

To isolate the contribution of each scaffolding component, we ran an ablation using only the correction node (without the summarization module) on the AWQ configuration with full 32K context. Table~\ref{tab:ablation} presents the results.

\begin{table}[ht]
\centering
\caption{Ablation study on AWQ configuration (32K context). ``Correction Only'' uses the isolated correction node without summarization.}
\label{tab:ablation}
\begin{tabular}{lccc}
\toprule
\textbf{Difficulty} & \textbf{Baseline} & \textbf{Correction Only} & \textbf{Full Scaffold} \\
\midrule
Aggregate & 3.0 & 4.8 & \textbf{5.9} \\
Difficulty 1 & 5.3 & 12.3 & \textbf{14.0} \\
Difficulty 2 & 2.1 & 2.1 & \textbf{4.2} \\
Difficulty 3 & 1.6 & 0.0 & 0.0 \\
\bottomrule
\end{tabular}
\end{table}

The correction node alone accounts for the majority of the aggregate improvement (3.0\%$\to$4.8\%), while the summarization module provides an additional boost (4.8\%$\to$5.9\%), particularly at difficulty-2 where it doubles performance from 2.1\% to 4.2\%. This is consistent with the hypothesis that the summarization module's primary value is in preserving critical state (authentication tokens, intermediate results) across longer interaction sequences---exactly the kind of information needed for multi-step difficulty-2 tasks.

Interestingly, the correction-only ablation shows 0.0\% on difficulty-3, compared to the baseline's 1.6\%. This confirms our earlier observation that history isolation can harm performance on complex tasks where the correction node lacks sufficient context to generate valid patches.

\subsection*{Failure Mode Shift: Before vs.\ After Scaffolding}

A key question is whether the scaffolding targets the failure modes it was designed to address. Table~\ref{tab:failure_shift} presents a direct comparison of failure mode distributions before and after scaffolding on the AWQ configuration. Because the scaffolded model solves more tasks (and thus has fewer failures), we report both raw counts and the proportion of total failures for each category.

\begin{table}[ht]
\centering
\caption{Failure mode comparison before and after scaffolding (AWQ, \texttt{test\_normal}). Counts reflect primary failure classification across all difficulty levels. The scaffolded model has fewer total failures (158 vs.\ 163) due to improved task goal completion.}
\label{tab:failure_shift}
\begin{tabular}{lcccc}
\toprule
 & \multicolumn{2}{c}{\textbf{Baseline (AWQ)}} & \multicolumn{2}{c}{\textbf{Scaffolded (AWQ)}} \\
\cmidrule(lr){2-3} \cmidrule(lr){4-5}
\textbf{Failure Category} & \textbf{Count} & \textbf{\%} & \textbf{Count} & \textbf{\%} \\
\midrule
Auth / credentials & 46 & 28.2 & 43 & 27.2 \\
Reasoning / planning & 43 & 26.4 & 54 & 34.2 \\
API params / schema & 29 & 17.8 & 15 & 9.5 \\
Other & 24 & 14.7 & 19 & 12.0 \\
Missing API / wrong name & 13 & 8.0 & 17 & 10.8 \\
Repetition / loop & 3 & 1.8 & 2 & 1.3 \\
Pagination / incomplete & 2 & 1.2 & 4 & 2.5 \\
Formatting / code block & 2 & 1.2 & 2 & 1.3 \\
Context length / token limit & 1 & 0.6 & 0 & 0.0 \\
Tooling runtime error & --- & --- & 2 & 1.3 \\
\bottomrule
\end{tabular}
\end{table}

Several shifts are notable. First, API parameter/schema mismatches drop sharply from 17.8\% to 9.5\% of failures, nearly halved, consistent with the correction node's access to API documentation enabling it to fix argument name and type errors. Second, repetition/loop failures decline from 1.8\% to 1.3\%, confirming that the correction node's history isolation helps break repetitive cycles. Third, context length failures are eliminated entirely, validating the summarization module's effectiveness. Fourth, and most importantly, reasoning/planning errors increase in proportion from 26.4\% to 34.2\%, not because the model reasons worse, but because the scaffolding resolves the mechanical failures that previously masked the model's fundamental planning limitations. This ``unmasking'' effect suggests that further gains for small models on AppWorld will require improvements to core reasoning capabilities, whether through training, more sophisticated prompting, or architectural changes.

Table~\ref{tab:post_scaffold_failures} provides a per-difficulty breakdown of the post-scaffolding failure modes for the AWQ model.

\begin{table}[ht]
\centering
\caption{Top failure modes by difficulty after full scaffolding (AWQ, 158 failed tasks).}
\label{tab:post_scaffold_failures}
\begin{tabular}{clc}
\toprule
\textbf{Difficulty} & \textbf{Failure Category} & \textbf{Count} \\
\midrule
\multirow{3}{*}{1} & Reasoning / planning & 18 \\
                    & Auth / credentials & 8 \\
                    & API params / schema & 8 \\
\midrule
\multirow{3}{*}{2} & Auth / credentials & 15 \\
                    & Reasoning / planning & 14 \\
                    & Missing API / wrong name & 8 \\
\midrule
\multirow{3}{*}{3} & Reasoning / planning & 22 \\
                    & Auth / credentials & 20 \\
                    & Other & 6 \\
\bottomrule
\end{tabular}
\end{table}

\section*{Discussion}

Our results demonstrate that structured inference-time scaffolding can meaningfully improve small language model agents on complex tool-use benchmarks without any additional training compute. The roughly 2$\times$ improvement across both quantized and full-precision configurations suggests that the scaffolding addresses systematic failure patterns. In the RL framing, the scaffolded policy $\pi^{\text{scaf}}$ achieves substantially higher expected reward $J(\pi^{\text{scaf}}) \gg J(\pi_\theta)$ despite sharing the same frozen parameters $\theta$, demonstrating that the effective policy can be significantly improved through input/output transformations alone, without policy gradient updates.

The failure mode analysis reveals an important insight about the nature of small model failures on agentic tasks. Unlike frontier models that primarily fail on complex reasoning, the 8B model exhibits a high proportion of ``mechanical'' failures, such as authentication handling, API schema compliance, and context management, that are amenable to inference-time correction. The before-and-after failure mode comparison (Table~\ref{tab:failure_shift}) provides direct evidence that the scaffolding selectively targets these mechanical errors: API schema mismatches are nearly halved, repetitive loops are reduced, and context length failures are eliminated entirely. Meanwhile, reasoning and planning errors increase as a proportion of remaining failures, revealing the model's core planning limitations that were previously obscured by more easily addressable errors. This ``unmasking'' effect suggests a natural progression for improving small model agents. First eliminate mechanical failures through scaffolding, then address residual reasoning limitations through training or architectural improvements.

The precision-versus-context tradeoff between FP16 and AWQ configurations merits further investigation. Despite the AWQ model having access to nearly 3$\times$ the context window, the FP16 model consistently outperforms it. In our formalism, this suggests that the quality of the per-step policy $\pi_\theta^{\text{agent}}(a_t \mid \tilde{h}_t)$ degrades more under quantization ($\theta \to \hat{\theta}$) than it gains from reduced partial observability (longer $\tilde{h}_t$), at least when $\pi_\theta^{\text{sum}}$ is employed to mitigate context loss. Distinguishing between these explanations would require an ablation that disables the summarization module in the AWQ configuration, isolating the contribution of extended context from the contribution of numerical precision. We leave this as a direction for future work.

The correction model's dual role, both as an error corrector and a loop breaker, proved effective for simpler tasks but introduced a tradeoff at higher difficulties. In RL terms, the regularization from restricting $\pi_\theta^{\text{correct}}$'s observation set to $(a_t, d_t)$ reduces variance from corrupted histories but introduces bias when the optimal correction requires long-horizon state information. Future work could explore \emph{selective history injection}, where only verified artifacts are passed to the correction model, expanding its observation set to $(a_t, d_t, \hat{v}_t)$ with validated state variables, or learned gating mechanisms that decide when history isolation is beneficial versus harmful.

More broadly, our results suggest that inference-time scaffolding and training-based optimization are complementary rather than competing strategies. The failure mode shift in Table~\ref{tab:failure_shift} indicates that once mechanical failures are addressed through scaffolding, the residual failure distribution is dominated by reasoning and planning errors that are unlikely to yield further to prompt engineering alone. This points toward a natural two-stage improvement path: apply inference-time scaffolding to eliminate addressable mechanical failures, then target the exposed reasoning limitations through RL fine-tuning or architectural improvements. Potentially applied to a model that is already scaffolded, compounding the gains from both approaches.


\subsection*{Limitations}
Our study has several limitations. First, we evaluate only on \texttt{test\_normal}; 
we did not attempt \texttt{test\_challenge} due to compute budget constraints and 
the historically near-zero performance of small models on that split, though future 
work with stronger base models may find it informative. Second, all results use 
single-trial greedy decoding (pass@1), which produces higher-variance estimates 
than multi-trial protocols on a benchmark of 168 tasks, the resulting confidence 
intervals are wide enough that our numerical improvement over DeepSeek-Coder 33B 
Instruct is not statistically significant at 95\%, and should be interpreted as 
a directional result rather than a definitive claim. Third, the failure mode 
classification relies on GPT-4o as an evaluator, which may introduce systematic 
biases in the taxonomy; human annotation of a subset would strengthen this analysis. 
Fourth, our compute budget limited hyperparameter tuning for the credential manager 
and summarization thresholds, and the credential manager in particular was abandoned 
at break-even performance, a more carefully tuned implementation may recover 
gains at higher difficulty levels. Finally, results are not directly comparable to concurrent work using multi-trial protocols such as pass@4 or Avg@4 \cite{cao2025reme}, which report substantially higher absolute 
numbers on the same split due to the evaluation methodology rather than underlying 
performance differences.

\section*{Conclusion}

We have shown that a three-tier inference scaffolding pipeline, where deploying the same frozen model as a main agent, a context summarizer, and an isolated corrector, can approximately double the performance of Qwen3-8B on the AppWorld benchmark using only a single 24\,GB GPU. Our approach requires no additional training and operates purely at test time, making it immediately applicable to any deployed small language model agent. The scaffolded 8B model achieves 8.9\% task goal completion on full-precision inference, surpassing DeepSeek-Coder 33B Instruct at 7.1\%, and demonstrating that intelligent allocation of inference-time compute can partially compensate for model scale in complex agentic settings.

\section*{Acknowledgements}

All experiments were conducted on a single NVIDIA GPU with 24\,GB VRAM. We thank the AppWorld team for their benchmark and open-source evaluation framework.

\section*{Author contributions statement}

A.A. conceived the experimental methodology, designed and implemented the inference scaffolding pipeline, conducted all experiments, and performed the failure mode analysis. B.B. provided research direction and contributed to manuscript preparation. All authors reviewed the manuscript.

\section*{Ethical Approval}
This study was reviewed by the Institutional Review Board of New York University and was determined to be exempt from further review (protocol number: IRB-FY2026-11424).

\section*{Competing interests}
The authors declare no competing interests.

\section*{Data Availability}
Code for the inference scaffolding pipeline is available at \url{https://github.com/Aimpoint-Digital/appworld-agent}.

\bibliography{references}

\end{document}